\documentclass[letterpaper,10pt,conference]{ieeeconf}
\IEEEoverridecommandlockouts
\usepackage{cite}
\usepackage{amsmath,amssymb,amsfonts}
\usepackage{graphicx}
\usepackage{booktabs}
\usepackage{hyperref}   
\usepackage{cleveref}
\usepackage[table]{xcolor}
\usepackage{subcaption}
\usepackage{pgfplots}
\pgfplotsset{compat=1.16}
\usepackage{algorithm}
\usepackage{algpseudocode}

\newtheorem{theorem}{Theorem}[section]
\newtheorem{lemma}[theorem]{Lemma}
\newtheorem{proposition}[theorem]{Proposition}

\newtheorem{assumption}[theorem]{Assumption}

\usepackage{mathtools}

\newcommand{\E}{\mathbb{E}}

\title{\LARGE\bfseries 
KKT-Perturbation-Aware Communication for Fully Distributed Online Convex Optimization with Nonseparable Coupling}

\author{
\authorblockN{Zhaoye Pan\textsuperscript{1}, Tao  Li\textsuperscript{1,\textdagger},~\IEEEmembership{Member, IEEE}, Yunjie Guo\textsuperscript{2}, and Guanpu Chen\textsuperscript{3},~\IEEEmembership{Member, IEEE}}
\thanks{\textsuperscript{1}Z. Pan and T. Li are with the Department of Systems Engineering, City University of Hong Kong, Hong Kong SAR, China (e-mail: iszhaoyepan@gmail.com, li.tao@cityu.edu.hk).
\textsuperscript{2}Y. Guo is with the State Grid Electric Power Research Institute, Nanjing, China (email: guoyunjie@sgepri.sgcc.com.cn).
\textsuperscript{3}G. Chen is with the School of Automation, Southeast University, Nanjing, China (email: guanpu\_chen@seu.edu.cn).
\textsuperscript{\textdagger}Corresponding author: T. Li. This work was supported in part by a CityU internal grant (Grant No. 9610792).
}
}

\begin{document}
\maketitle

\begin{abstract}
Nonseparable costs and coupled constraints in distributed online convex optimization often arise when system performance and operational constraints depend on the joint configuration of all agents and cannot be decomposed into individual contributions. Such nonseparability leads to fully distributed decision-making with challenging partial local information, where each agent evaluates its local primal–dual update using estimated global information via a local belief of the joint action and the shared Lagrange multiplier to be exchanged with neighbors. While the belief consensus established by prior works ensures no-regret learning, exchanging beliefs comprising full-dimensional primal and dual states at every round imposes a communication overhead that scales with both the joint decision dimension and the network size. Existing event-triggered schemes typically activate communication according to state variation or consensus error. Under nonseparable coupling, however, errors of equal magnitude can produce markedly different perturbations after passing through the KKT operator. 
We therefore propose a randomized KKT-perturbation-aware communication mechanism, where the triggering of each agent's primal and dual communication channels depends on their impact on the KKT operator. The proposed mechanism transfers the adaptive-sampling principle of coordinate descent from allocating computation across decision blocks to allocating communication across information channels: channels with larger estimated perturbations receive greater priority for refresh. Under standard regularity assumptions, the resulting algorithm achieves \(O({T}^{2/3})\) expected static regret and \(O({T}^{2/3})\) expected cumulative constraint violation.
Numerical experiments demonstrate improved learning and constraint-enforcement performance per transmitted bit over belief consensus baselines.
\end{abstract}


\section{Introduction}
Distributed online convex optimization (D-OCO) provides
a standard framework for multi-agent sequential decision-making under
uncertainty, with applications spanning cyber-physical network systems~\cite {tao22confluence}. In D-OCO with
time-varying coupled constraints, a network of agents repeatedly selects
local actions, after which a global cost and a vector of shared constraint
functions are revealed, and the agents seek to minimize the cumulative cost
while keeping the long-term constraint violation growth sublinear
\cite{yi19doco,yuan20ma-coco,Tao_blackwell,sinha24opt-coco}. Most existing formulations assume that
both the objective and the coupled constraints are \emph{separable}, i.e.,
expressed as sums of agent-specific local functions
\cite{yuan20ma-coco}. Under separability, each agent's partial gradient
of the global objective coincides with the gradient of its local cost, so
that gradient-based primal--dual updates can be assembled from local
information and neighbor communication.

The separability assumption, however, fails to capture team-based
decision-making problems in which system performance and safety depend on
the \emph{joint} configuration of all agents and cannot be decomposed into
agent-specific contributions. A concrete instance is the multi-camera cooperative traffic
sensing studied in~\cite{tao24picol,tao25dima}, where networked pan-tilt cameras
are tilted to different road segments to maximize urban-level traffic volume
coverage. Such a sensing metric cannot be decomposed into camera-specific objectives. Essentially, nonseparability arises from
the agent-agnostic and symmetric structure of large-population systems
\cite{pan-tao22noneq,pan-tao23delay} and from distributed equilibrium seeking in games~\cite{raginsky16nonseparable,pavel20distributed-gne,guanze21local-gne,liu23game-erm,liao2026fully}.



Nonseparability leads to a challenging information structure for
multi-agent decision making \cite{tao22info}, referred to as fully distributed optimization under partial-decision information \cite{pavel20distributed-gne}. Each agent can access only the decisions of its neighbors, from which the global
quantities of interest must be estimated. A common estimation scheme shared by prior works on fully distributed optimization relies on \textit{belief consensus}, where each agent maintains and exchanges their local estimates of joint actions and shared Lagrangian multipliers 
\cite{raginsky16nonseparable,pavel20distributed-gne,liao2026fully,pan2026distributed}. The price to pay is the increasing communication
overhead with respect to the number of agents and action space dimension. 

To address the communication overhead, there exist two lines of work on communication-efficient mechanisms; yet, neither is
well aligned with the nonseparable setting. Event-triggered communication
\cite{yang2026dynamic,chen2025event,ding2024distributed} activates transmission according to the \textit{state variation}, the
discrepancy between the current state and its last transmitted or cached
value, possibly with dynamic, adaptive, or switching thresholds. The triggering
threshold design refines the \emph{timing} of communication, yet the
underlying communication priority remains governed by state variation.
Compressed communication methods \cite{liu2023communication,song2022compressed,zhu2026distributed} reduce the
payload of each message, typically under iteration-wise or prescribed
transmission schedules, and do not address \emph{when} a channel is worth
activating. 

Our key observation is that, under nonseparable coupling,
\emph{state variation is not an adequate measure of the impact of
stale information}: two state errors of equal magnitude can produce
markedly different perturbations once passed through the KKT operator,
because the agent's local update depends on the global gradient, the constraint
Jacobian, and the current multiplier. A barely changing belief may
require no transmission, whereas a small dual variable drift that shifts
the constraint-price term in the primal direction must be refreshed
promptly. Triggers based on  state variation therefore may misallocate
communication resources under nonseparable coupling.

We propose a
\emph{KKT-perturbation-aware} communication mechanism for fully distributed optimization under time-varying, nonseparable costs and constraints. Each agent possesses two communication channels: a primal-belief channel and a dual channel. We define its \emph{impact} as the
perturbation that retaining the stale cached value would induce on
the local KKT operator surrogate using the revealed feedback. The perturbation-based impact is then mapped to channel activation probabilities, which are closely connected to coordinate-sampling probabilities in randomized coordinate descent
\cite{perekrestenko2017faster,leluc2022sgd}.


Our contributions are summarized as follows.
\begin{itemize}
    \item \emph{KKT-perturbation-aware communication:}
    To account for the nonseparability in distributed online convex optimization, we measure the impact of each agent's communication channels on the KKT operator by quantifying the perturbation induced by stale, cached primal and dual states, and we map this perturbation to adaptive channel activation sampling to reduce communication. 

    \item \emph{Optimization-theoretic interpretation:} We connect the proposed mechanism to adaptive sampling in randomized coordinate descent, and show in Prop.~\ref{prop:ordering} that the activation probabilities preserve the channel-priority ordering induced by a centralized communication-allocation problem.
    \item \emph{Regret guarantees:}
    Under standard regularity and a uniform Slater condition, the KKT perturbation-aware algorithm achieves $O({T}^{2/3})$ expected static regret and $O({T}^{2/3})$ expected cumulative constraint violation (CCV) (Thm.~\ref{thm:regret} and Thm.~\ref{thm:ccv}).
    \item \emph{Numerical validation:}
    Experiments on a distributed quadratic program with dense coupling show that, under a matched communication budget, the proposed mechanism consistently incurs smaller cumulative regret than the baseline belief-consensus algorithms across different network sizes.
\end{itemize}

\noindent\textbf{Related works.} Distributed online optimization with
time-varying coupled constraints has been studied under separable
objectives and constraints in \cite{yi19doco,yuan20ma-coco}, with bandit feedback
in \cite{yangtao20doco-bandit}, and the optimal $O(\sqrt{T})$ regret--cumulative constraint violation (CCV) trade-off
for the single-agent case was settled in \cite{sinha24opt-coco}. Nonseparable
global objectives in distributed (offline) optimization were addressed via
coordinate descent \cite{lina13state_game} and coordinate dual averaging
\cite{raginsky16nonseparable}, both requiring agents to track estimates of the joint
decision. \cite{pan2026distributed} extended this belief-sharing idea to the online
constrained setting but assumed per-round broadcasting of full states.
Event-triggered Nash equilibrium seeking under state-variation triggers was
developed in \cite{yang2026dynamic,chen2025event,ding2024distributed}, and compressed distributed
optimization in \cite{liu2023communication,song2022compressed,zhu2026distributed}. To our knowledge, no prior
work allocates communication according to the estimated perturbation of the
primal--dual (KKT) operator, which is precisely the quantity that stale
information corrupts in the nonseparable regime.

\section{Problem formulation}\label{sec:problem}
Consider $N$ agents indexed by $\mathcal{V}=\{1,\ldots,N\}$, networked
by an undirected graph $\mathcal{G}=(\mathcal{V},\mathcal{E})$. At each
round $t\in\mathbb{N}$, agent $i\in\mathcal{V}$ selects a local action
$a_{i,t}\in\mathcal{A}_i\subseteq\mathbb{R}^{d_i}$, where
$\mathcal{A}_i$ is convex and compact. The joint action is
$a_t\coloneqq\operatorname{col}(a_{1,t},\ldots,a_{N,t})\in
\mathcal{A}\coloneqq\prod_{i=1}^N\mathcal{A}_i\subseteq\mathbb{R}^{d}$
with $d\coloneqq\sum_{i=1}^N d_i$. After $a_t$ is implemented, the
environment reveals a global, nonseparable cost $f_t:\mathcal{A}
\rightarrow\mathbb{R}$ and a vector of $m$ nonseparable, convex
constraint functions $g_t:\mathcal{A}\rightarrow\mathbb{R}^m$, whose
$k$-th component is denoted by $g_{k,t}$. The agents seek to generate
$\{a_t\}_{t=1}^T$ that minimizes the cumulative cost subject to the
constraints in the long run:
\begin{equation*}
    \min_{\{a_t\}_{t=1}^T\subseteq\mathcal{A}}
    \ \sum_{t=1}^{T}f_t(a_t)
    \quad
    \text{s.t.}\;\;
    g_t(a_t)\preceq\mathbf{0},
    \quad
    \forall t\in\{1,\ldots,T\},
    \label{eq:online_problem}
\end{equation*}
where $\preceq$ is  componentwise. We evaluate an online
distributed algorithm by the static regret
\begin{equation*}
    \operatorname{Reg}(T)
    \coloneqq
    \sum_{t=1}^{T}
    \left[
        f_t(a_t)-f_t(a^\star)
    \right],
    \label{eq:static_regret}
\end{equation*}
where
$a^\star\in\arg\min_{a\in\mathcal{A}}\{\sum_{t=1}^T f_t(a):
g_t(a)\preceq\mathbf{0},\ \forall t\}$
is the best fixed feasible joint action in hindsight, and by the
componentwise cumulative constraint violation
\begin{equation*}
    \operatorname{CCV}_k(T)
    \coloneqq
    \left[
        \sum_{t=1}^{T}g_{k,t}(a_t)
    \right]_+,
    \qquad
    k\in\{1,\ldots,m\},
    \label{eq:ccv}
\end{equation*}
where $[z]_+\coloneqq\max\{0,z\}$. Our standing assumptions are
standard in distributed online constrained optimization
\cite{yuan20ma-coco}.

\begin{assumption}\label{ass:regularity} The following regularity and constraint qualification conditions hold.
\textup{(i)} For every $t$, the functions $f_t$ and
$g_{k,t}$, $k\in\{1,\ldots,m\}$, are convex and continuously
differentiable on $\mathcal A$. The set
$
\mathcal A=\prod_{i=1}^N\mathcal A_i
$
is compact and convex, with diameter $D_{\mathcal A}$. There exist
constants $G_f,\;G_g,\;G_H,\;L_f,\;L_g,\;L_{\nabla g}>0,$
such that, for all $x,x'\in\mathcal A$ and all $t$,
\begin{align*}
&\|\nabla f_t(x)\| \le G_f,
\|\nabla g_t(x)\|_{\mathrm F} \le G_g,
\|g_t(x)\| \le G_H,\\
&\|\nabla f_t(x)-\nabla f_t(x')\|
\le L_f\|x-x'\|,\\
&\|g_t(x)-g_t(x')\|
\le L_g\|x-x'\|,\\
&\|\nabla g_t(x)-\nabla g_t(x')\|_{\mathrm F}
\le L_{\nabla g}\|x-x'\|.
\end{align*}

\textup{(ii)} A uniform Slater condition holds: there exist
$\bar a\in\mathcal A$ and $\varepsilon_{\mathrm S}>0$ such that
$
g_t(\bar a)\preceq -\varepsilon_{\mathrm S}\mathbf 1_m,\forall t.
$
The dual iterates are projected onto
$
\Lambda\coloneqq[0,\Lambda_{\max}]^m,
$
where $\Lambda_{\max}$ is chosen sufficiently large so that the projection
does not exclude the optimal multipliers under the uniform Slater
condition \cite{yi19doco}, and \cite[Prop.~1]{pan2026distributed} presents a characterization of such $\Lambda_{\max}$.

\textup{(iii)} The graph $\mathcal G$ is connected and undirected.
The mixing matrix $W=[w_{ij}]$ is symmetric, doubly stochastic, and
compatible with $\mathcal G$. Its second-largest eigenvalue modulus
satisfies
$\sigma\coloneqq
\max\{|\lambda_2(W)|,\ldots,|\lambda_N(W)|\}\in[0,1).$ The proposed framework can be generalized to directed graphs by replacing the doubly stochastic belief-consensus step with a push-sum-type mechanism \cite{raginsky16nonseparable,lin2025analysis}.
\end{assumption}

\paragraph*{Notation}
$\Pi_{\mathcal S}$ denotes the Euclidean projection onto a
closed convex set $\mathcal S$, $\|\cdot\|$ the Euclidean norm,
$\|\cdot\|_{\mathrm F}$ the Frobenius norm, $I_d$ the $d\times d$
identity, $\otimes$ the Kronecker product, and
$\mathsf{R}_i=[\mathbf{0}, I_{d_i}, \mathbf{0}]\in\mathbb{R}^{d_i\times d}$ the block-selection
matrix with $\mathsf{R}_ix=x_i$, whose transpose
$\mathsf{U}_i\coloneqq\mathsf{R}_i^{\top}$ injects a block $x_i$ back
into $\mathbb{R}^d$ with all other entries being zeros. Similarly, we have $\mathsf{R}_{-i}=\operatorname{col}(\mathsf{R}_{1},\ldots, \mathsf{R}_{i-1},\mathsf{R}_{i+1}, \ldots, \mathsf{R}_{N})\in\mathbb{R}^{(d-d_i)\times d}$, and $\mathsf{U}_{-i}\coloneqq\mathsf{R}_{-i}^{\top}=[\mathsf{U}_1, \ldots, \mathsf{U}_{i-1}, \mathsf{U}_{i+1}, \ldots, \mathsf{U}_{N}]$.  $\mathcal{N}_i$ denotes the set of neighboring nodes of $i$. Each agent
communicates over two channels indexed by
$c\in\{z,\lambda\}$, carrying the primal belief and the dual
variable, and we write $x_{i,t}^{z}\coloneqq z_{i,t}$ and
$x_{i,t}^{\lambda}\coloneqq\lambda_{i,t}$ for the channel
states. Stacked vectors are denoted by dropping the agent
index, e.g.,
$Z_t\coloneqq\operatorname{col}(z_{1,t},\ldots,z_{N,t})$.

\section{Perturbation-Aware Communication}\label{sec:algorithm}
Our starting point is the full-communication belief-consensus in \cite{pan2026distributed}, which we briefly review in the following. Since $f_t$ and $g_t$ are
nonseparable, agent $i$ cannot evaluate its Lagrangian partial
gradient from its own action alone and hence maintains a
\emph{belief} of the joint action,
\begin{equation*}
    z_{i,t}
    \coloneqq
    \operatorname{col}
    \bigl(
        z_{i,t}^{1},\ldots,z_{i,t}^{N}
    \bigr)
    \in\mathcal{A},
    \qquad
    z_{i,t}^{i}=a_{i,t},
    \label{eq:belief}
\end{equation*}
where the block $z_{i,t}^i$ is the implemented action and the remaining blocks are the
estimates of other agents' actions, together with a local copy
$\lambda_{i,t}\succeq\mathbf0$ of the shared multiplier. All agents exchange their
current $(z_{i,t},\lambda_{i,t})$ at every round. 

To reduce the communication overhead, we here equip
each agent with two communication channels: a \emph{primal-belief
channel} carrying $z_{i,t}$ and a \emph{dual channel} carrying
$\lambda_{i,t}$. Unlike prior works \cite{pavel20distributed-gne,liao2026fully, pan2026distributed}, we allow each channel to remain silent for multiple
rounds. During the online decision-making, agent $i$ keeps a sender-side reference
$s_{i,t}^{z}\in\mathcal{A}$ and $s_{i,t}^{\lambda}\succeq\mathbf0$,
equal to its most recently transmitted states, while each neighbor
$j\in\mathcal{N}_i$ keeps receiver-side caches
$\widehat z_{j|i,t}\in\mathcal{A}$ and
$\widehat\lambda_{j|i,t}\succeq\mathbf0$. The two-sided states are
initialized consistently, $\widehat z_{j|i,1}=s_{j,1}^{z}$ and
$\widehat\lambda_{j|i,1}=s_{j,1}^{\lambda}$, and perform updates: for $c\in \{z, \lambda\}$,
\begin{subequations}
\begin{align}
    s_{i,t+1}^{c}
    &=
    \xi_{i,t}^{c}\,x_{i,t}^{c}
    +
    \bigl(1-\xi_{i,t}^{c}\bigr)s_{i,t}^{c},
    \label{eq:s-cache-update} \\
    \widehat x_{j|i,t+1}^{c}
    &=
    \xi_{j,t}^{c}\,x_{j,t}^{c}
    +
    \bigl(1-\xi_{j,t}^{c}\bigr)\widehat x_{j|i,t}^{c},
    \label{eq:r-cache_update}
\end{align}    
\label{eq:cache_update}
\end{subequations}
where $x_{i,t}^{z}\coloneqq z_{i,t}$, $x_{i,t}^{\lambda}\coloneqq
\lambda_{i,t}$, and $\xi_{i,t}^{z},\xi_{i,t}^{\lambda}\in\{0,1\}$ are
the channel activation indicators: when $\xi_{i,t}^{c}=1$, agent $i$
broadcasts its current state on channel $c$ and both sender references
and receiver caches are refreshed; when $\xi_{i,t}^{c}=0$, nothing is
transmitted and all parties retain the stale value. By induction,
\eqref{eq:cache_update} preserves
$\widehat x_{j|i,t}^{c}=s_{j,t}^{c}$ for every receiver, so a channel
is stale \emph{by exactly the same amount} at the sender and at all of
its neighbors.

The communicated information enters the online optimization procedure only through
the \emph{cache-fused} consensus states
\begin{subequations}
\label{eq:fusion}
\begin{align}
    &\widetilde z_{i,t}
    \coloneqq
    (1-w_{ii})\,\mathsf{U}_i a_{i,t}
    +
    w_{ii}\,z_{i,t}
    +
    \sum_{j\in\mathcal{N}_i}
    w_{ij}\,
    \mathsf{U}_{-i}
        [\widehat z_{j|i,t+1}]_{-i},
    \label{eq:primal_fusion}
    \\
    &\widetilde\lambda_{i,t}
    \coloneqq
    w_{ii}\lambda_{i,t}
    +
    \sum_{j\in\mathcal{N}_i}
    w_{ij}\,\widehat\lambda_{j|i,t+1},
    \label{eq:dual_fusion}
\end{align}
\end{subequations}
i.e., the self-block of $\widetilde z_{i,t}$ is overridden by the
implemented action, $[\widetilde z_{i,t}]_i=a_{i,t}$, and only the
other agents' blocks are fused from the caches. Here,
$\mathsf{U}_{-i}$ embeds a block vector of dimension $d-d_i$ into
$\mathbb R^{d}$ with a zero $i$-th block. Because every term on the
right-hand side of \eqref{eq:fusion} is a previously implemented
feasible state, $\widetilde z_{i,t}\in\mathcal{A}$ and
$\widetilde\lambda_{i,t}\succeq\mathbf0$ hold naturally, and no projection is required. After $a_t$ is implemented
and $(f_t,g_t)$ is revealed, agent $i$ evaluates its local primal
direction at the fused state,
\begin{equation}
    d_{i,t}
    \coloneqq
    \nabla_{a_i}f_t(\widetilde z_{i,t})
    +
    \nabla_{a_i}g_t(\widetilde z_{i,t})^{\!\top}
    \widetilde\lambda_{i,t},
    \label{eq:local_direction}
\end{equation}
and performs the projected blockwise updates
\begin{subequations}
\label{eq:updates}
\begin{align}
    a_{i,t+1}
    &=
    \Pi_{\mathcal{A}_i}
    \Bigl[
        a_{i,t}-\alpha_t d_{i,t}
    \Bigr],
    \label{eq:primal_update}
    \\
    z_{i,t+1}
    &=
    \widetilde z_{i,t}
    +
    \mathsf{U}_i
    \bigl(
        a_{i,t+1}-a_{i,t}
    \bigr),
    \label{eq:belief_update}
    \\
    \lambda_{i,t+1}
    &=\Pi_{\Lambda}\Bigl[
        \lambda_{i,t}
        +
        \beta_t g_t(\widetilde z_{i,t})
    \Bigr],
    \label{eq:dual_update}
\end{align}
\label{eq:communication-primal-dual}
\end{subequations}
where $\{\alpha_t\}$ and $\{\beta_t\}$ are primal and dual
stepsizes and $\mathsf{U}_i$ injects the updated self-block into the
belief while maintaining the fused estimates of all other blocks.
Thus, relative to \cite{pan2026distributed}, the primal--dual
update remains largely unchanged except that exact neighbor states are replaced
by cached ones. When
$\xi_{i,t}^{z}=\xi_{i,t}^{\lambda}=1$ for all $i$ and $t$, the update \eqref{eq:communication-primal-dual} reduces to the full-communication belief consensus \cite{pavel20distributed-gne,pan2026distributed}.

\paragraph*{KKT-perturbation triggering}
Key to our communication mechanism is the \emph{impact} of a stale channel, which points to the perturbation it induces on the primal--dual operator.
Define the smooth forward part of the stage-$t$ KKT operator as
$\mathcal{B}_t(x,\lambda)\coloneqq
\operatorname{col}\bigl(\nabla f_t(x)+\nabla g_t(x)^{\!\top}\lambda,\,
-g_t(x)\bigr)$. Agent $i$ only requires its own
primal block, and therefore operates on the local component
\begin{equation}
    \mathcal{B}_{i,t}^{\mathrm{loc}}(x,\lambda)
    \coloneqq
    \operatorname{col}
    \Bigl(
        \nabla_{a_i}f_t(x)
        +
        \nabla_{a_i}g_t(x)^{\!\top}\lambda,\,
        -g_t(x)
    \Bigr).
    \label{eq:local_operator}
\end{equation}
Errors of equal magnitude can induce markedly different perturbations
once passed through \eqref{eq:local_operator}, and the two channels
fail in characteristically different ways. A stale primal belief
$x=s_{i,t}^{z}$ distorts three terms of
$\mathcal{B}_{i,t}^{\mathrm{loc}}$ simultaneously: the objective
gradient $\nabla_{a_i}f_t$, the multiplier-weighted constraint
gradient $\nabla_{a_i}g_t^{\top}\lambda$, and the constraint residual
$g_t$ that drives the dual update. A stale multiplier
$\lambda=s_{i,t}^{\lambda}$, by contrast, perturbs the primal direction
through a single channel, namely the correction
$\nabla_{a_i}g_t^{\top}(\lambda_{i,t}-s_{i,t}^{\lambda})$, and leaves
the dual residual untouched. Most prior works use state-variation triggers, i.e., the discrepancies $\|z_{i,t}-s_{i,t}^{z}\|$ and
$\|\lambda_{i,t}-s_{i,t}^{\lambda}\|$ as the triggering indicator, and are blind to this asymmetry. In contrast, our insight is that the two channels must be triggered separately, by their distinct impacts on the operator.

Specifically, since the stage-$t$ operator is unavailable before
communication, each agent $i$ estimates its channel impact at the beginning of
round $t$ using the most recently revealed feedback $(f_{t-1},g_{t-1})$ as a surrogate:
\begin{subequations}
\label{eq:scores}
\begin{align}
    \Delta_{i,t}^{z}
    &\coloneqq
    \bigl\|
        \mathcal{B}_{i,t-1}^{\mathrm{loc}}
        (z_{i,t},\lambda_{i,t})
        -
        \mathcal{B}_{i,t-1}^{\mathrm{loc}}
        (s_{i,t}^{z},\lambda_{i,t})
    \bigr\|,
    \label{eq:primal_score}
    \\
    \Delta_{i,t}^{\lambda}
    &\coloneqq
    \bigl\|
        \mathcal{B}_{i,t-1}^{\mathrm{loc}}
        (z_{i,t},\lambda_{i,t})
        -
        \mathcal{B}_{i,t-1}^{\mathrm{loc}}
        (z_{i,t},s_{i,t}^{\lambda})
    \bigr\|.
    \label{eq:dual_score}
\end{align}
\end{subequations}
Both impact scores are locally accessible from
\eqref{eq:local_direction}, and constitute a one-step-delayed
surrogate for the operator perturbation at time $t$. By
Assumption~\ref{ass:regularity}, they are bounded by the state variation as
\begin{align*}
    \Delta_{i,t}^{z}
    &\le
    \bigl(
        L_f+L_{\nabla g}\,\|\lambda_{i,t}\|
    \bigr)
    \|z_{i,t}-s_{i,t}^{z}\|
    +
    L_g\|z_{i,t}-s_{i,t}^{z}\|,
    \nonumber \\
    \Delta_{i,t}^{\lambda}
    &\le
    G_g\|\lambda_{i,t}-s_{i,t}^{\lambda}\|.
    \label{eq:score_bounds}
\end{align*}

Each score is mapped to an activation probability by
\begin{equation}
    p_{i,t}^{c}
    \coloneqq
    p_{\min}^{c}
    +
    \bigl(1-p_{\min}^{c}\bigr)
    \varphi_c\!\left(
        \frac{\Delta_{i,t}^{c}}{\tau_{i,t}^{c}}
    \right),
    \qquad
    c\in\{z,\lambda\},
    \label{eq:probabilities}
\end{equation}
where $\varphi_c:\mathbb{R}_+\rightarrow[0,1]$ is continuous and
nondecreasing with $\varphi_c(0)=0$ and
$\lim_{u\rightarrow\infty}\varphi_c(u)=1$, $\tau_{i,t}^{c}>0$ is a scaling factor. $p_{\min}^{c}\in(p_0,1)$ is an activation
floor, with a threshold that guarantees that no channel remains indefinitely idle.



To sum up, let $\mathcal{F}_{t}^{-}$ be the $\sigma$-algebra generated
by all states, feedback, and communication randomness available
immediately before the round-$t$ triggering decision,
\begin{equation*}
    \mathbb{P}
    \bigl(
        \xi_{i,t}^{c}=1
        \,\big|\,
        \mathcal{F}_{t}^{-}
    \bigr)
    =
    p_{i,t}^{c},
    \qquad
    c\in\{z,\lambda\},
    \label{eq:bernoulli}
\end{equation*}
with $\xi_{i,1}^{z}=\xi_{i,1}^{\lambda}=1$ initially.

\paragraph*{Connection to Coordinate Descent}
The triggering rule \eqref{eq:probabilities} admits an
optimization-theoretic interpretation that distinguishes it from
state-variation-based scheduling: activation probabilities enter the
expected algorithm dynamics as explicit allocation variables, much as
adaptive sampling in randomized coordinate
descent, where they appear in the conditional expected-progress
bound and thereby govern the allocation of limited update opportunities
across coordinates \cite{perekrestenko2017faster,leluc2022sgd}. Here,
the scarce resource is not computation but communication: every agent
performs its local primal--dual update at every round, while communication activations are limited and must be allocated across the $2N$
primal-belief and dual channels.

To expose this connection, stack the post-decision staleness
perturbations of both channels of agent $i$ as
\begin{align}
    \varepsilon_{i,t}
    &\coloneqq
    \operatorname{col}
    \Bigl(
        \bigl(1-\xi_{i,t}^{z}\bigr)
        \bigl[
            \mathcal{B}_{i,t-1}^{\mathrm{loc}}(z_{i,t},\lambda_{i,t})
            -
            \mathcal{B}_{i,t-1}^{\mathrm{loc}}(s_{i,t}^{z},\lambda_{i,t})
        \bigr],
        \, \nonumber \\
        &\bigl(1-\xi_{i,t}^{\lambda}\bigr)
        \bigl[
            \mathcal{B}_{i,t-1}^{\mathrm{loc}}(z_{i,t},\lambda_{i,t})
            -
            \mathcal{B}_{i,t-1}^{\mathrm{loc}}(z_{i,t},s_{i,t}^{\lambda})
\bigr]\Bigr).\label{eq:stale_channel_perturbation}
\end{align}
Conditional on $\mathcal{F}_{t}^{-}$, both scores in
\eqref{eq:scores} are fixed and the two indicators are independent
Bernoulli variables, so the conditional second moment of
\eqref{eq:stale_channel_perturbation} decomposes exactly as
\begin{equation}
    \mathbb{E}
    \bigl[
        \|\varepsilon_{i,t}\|^2
        \,\big|\,
        \mathcal{F}_{t}^{-}
    \bigr]
    =
    \bigl(1-p_{i,t}^{z}\bigr)
    \bigl(\Delta_{i,t}^{z}\bigr)^2
    +
    \bigl(1-p_{i,t}^{\lambda}\bigr)
    \bigl(\Delta_{i,t}^{\lambda}\bigr)^2.
    \label{eq:conditional_perturbation_identity}
\end{equation}
Increasing the activation probability of a channel reduces its expected
contribution to the operator perturbation at the marginal rate given by
its squared score. Hence, \eqref{eq:conditional_perturbation_identity} is an analogue of the conditional expected-descent bound in coordinate sampling, with the
probabilities $p_{i,t}^{c}$ appearing as design variables rather than
heuristic thresholds.

This observation suggests an idealized benchmark for the allocation of
communication. Index all channels by
$k\in\mathcal{C}\coloneqq\{(i,z),(i,\lambda):i\in\mathcal{V}\}$, and let $\gamma_{k,t}\geq 0$ denote the problem-specific weight associated with each channel, e.g., depending on connectivity and Lipschitz constants. Define the weighted impact
$w_{k,t}\coloneqq\gamma_{k,t}(\Delta_{k,t})^2$, the per-activation
transmission cost $c_{k,t}>0$, and an expected communication budget
$B_t$. If channel $k$ is not activated, its expected unresolved
contribution to the operator perturbation is $(1-p_{k,t})w_{k,t}$;
allocating a limited budget so as to minimize the total expected
perturbation leads to the fractional-knapsack problem
\begin{equation*}
    \min_{\{p_{k,t}\in[0,1]\}_{k\in\mathcal{C}}}
    \
    \sum_{k\in\mathcal{C}}
    (1-p_{k,t})\,w_{k,t}
    \quad
    \text{s.t.}\;\;
    \sum_{k\in\mathcal{C}}
    c_{k,t}\,p_{k,t}
    \leq B_t .
    \label{eq:ideal_allocation}
\end{equation*}
Its solutions prioritize channels by the impact-per-cost ratio
$w_{k,t}/c_{k,t}$ and exhibit a bang--bang triggering: there exists a $\mu_t$ such that the channel $k$ is activated when $w_{k,t}/c_{k,t}\geq\mu_t$; otherwise remains idle.
The linear programming above thus
identifies the impact-aware \emph{priority ordering} over channels, and
the threshold $\mu_t$ prices the shared budget.

The proposed rule \eqref{eq:probabilities} implements this priority
principle in a decentralized manner, in place of
the bang--bang control, which would require centralized knowledge of
all scores and of the budget. The following proposition assures that our perturbation-aware triggering preserves the priority established by the linear programming.

\begin{proposition}[Priority-ordering consistency]\label{prop:ordering}
Fix a round $t$ and suppose that (i) all channels share the same
transmission cost $c_{k,t}\equiv c_t$ and the same analytical weight
$\gamma_{k,t}\equiv\gamma_t$, and (ii) the triggering rule uses a
common strictly increasing link function $\varphi$, impact scale
$\tau_t$, and activation floor $p_{\min}$. Then for any two channels
$k,\ell\in\mathcal{C}$,
\begin{equation}
    \Delta_{k,t}>\Delta_{\ell,t}
    \quad\Longrightarrow\quad
    p_{k,t}>p_{\ell,t},
    \label{eq:ordering_consistency}
\end{equation}
and the ordering of the activation probabilities coincides with the
priority ordering induced by \eqref{eq:ideal_allocation}.
\end{proposition}
\begin{proof}
Under (i), $w_{k,t}>w_{\ell,t}$ if and only if
$\Delta_{k,t}>\Delta_{\ell,t}$, so both orderings are governed by the
scores. Under (ii), $p_{k,t}=\psi(\Delta_{k,t})$ with
$\psi(\Delta)\coloneqq
p_{\min}+(1-p_{\min})\varphi(\Delta/\tau_t)$ strictly increasing in
$\Delta$, hence $\Delta_{k,t}>\Delta_{\ell,t}$ implies
$p_{k,t}>p_{\ell,t}$.
\end{proof}

\section{Regret and Constraint Violation Analysis}\label{sec:analysis}
Throughout, $\mathbb{E}[\cdot]$ denotes expectation over the
communication randomness $\{\xi_{i,t}^{c}\}$. We analyze
\eqref{eq:fusion}--\eqref{eq:updates} with the dual step \eqref{eq:dual_update}.
All agents are initialized at a commonly known $a_{1}\in\mathcal A$,
\begin{equation}
z_{i,1}=s^{z}_{i,1}=a_{1},\qquad
\lambda_{i,1}=s^{\lambda}_{i,1}=0,\qquad i\in\mathcal V,
\end{equation}
with consistent receiver caches. We require the conditional
activation probability of every channel to be at least
$p_{\min}^{c}\ge p_{0}>0$ for some fixed $p_{0}\in(0,1)$. We use
horizon-dependent constant step sizes
\[
\alpha=\alpha_0T^{-2/3},\qquad
\beta=\beta_0T^{-2/3},\qquad \alpha_0,\beta_0>0.
\]
Let $B=\Lambda_{\max}$, $M_B=\sqrt m\,B$, $G_B=\sqrt N\,(G_f+G_gM_B)$, and
$\bar\lambda_t=N^{-1}\sum_i\lambda_{i,t}$. Feasibility and
projection give $\|d_t\|\le G_B$,
$\|a_{t+1}-a_t\|\le\alpha G_B$, and
$\|\lambda_{i,t+1}-\lambda_{i,t}\|\le\beta G_H$. We measure the
error of the fused information by
\[
e_t^2=\sum_i\|\widetilde z_{i,t}-a_t\|^2,\qquad
h_t^2=\sum_i\|\widetilde\lambda_{i,t}-\bar\lambda_t\|^2,
\]
which are referred to as the belief-to-action error and the multiplier disagreement,
respectively.

We first bound the accumulated communication errors.
\begin{lemma}\label{lem:staleness}
Under Assumption~\ref{ass:regularity}, the activation floors and
initialization, for any constant
stepsizes $\alpha,\beta>0$ and fixed $B$,
\begin{align*}
&\E\bigg[\sum_{t=1}^T e_t^2\bigg]\le CT\alpha^2,\ \sum_{t=1}^T\E [h_t]\le C(\alpha\beta T^2+T\beta),
\\
&\sum_{t=1}^T\E [e_t]\le CT\alpha,\
\E[\max_{t\le T}h_t^2]\le C(\alpha^2\beta^2T^2+T\beta^2).
\end{align*}
\end{lemma}

The following lemma characterizes the error's instantaneous effect on the primal--dual dynamics by a one-step
inexact inequality.

\begin{lemma}\label{lem:drift}
For all $a\in\mathcal A$ and $\nu\in\Lambda$, we have
\begin{align*}
&\mathcal L_t(a_t,\bar\lambda_t)-\mathcal L_t(a,\bar\lambda_t)
+\langle g_t(a_t),\nu-\bar\lambda_t\rangle\nonumber\le
\frac{V_t(\nu)-V_{t+1}(\nu)}{2\beta} \\
&+\frac{\|a_t-a\|^2-\|a_{t+1}-a\|^2}{2\alpha}
+\frac{\alpha G_B^2+\beta G_H^2}{2}
+R_t(a,\nu),
\end{align*}
where 
\begin{align*}
&  L_t(a,\nu)=f_t(a)+\langle\nu,g_t(a)\rangle,V_t(\nu)=N^{-1}\sum_i\|\lambda_{i,t}-\nu\|^2,\\
&R_t(a,\nu)=\langle \zeta_t ,a-a_t\rangle+\frac{1}{N}\sum_{i}\langle g_{i,t}-g_t(a_t), \lambda_{i,t}-\nu \rangle,\\
&\zeta_t=d_t-\nabla f_t(a_t)-\nabla g_t(a_t)^{\!\top}\bar{\lambda}_t.
\end{align*}
The above three terms arise from the inexact fused information, and in
particular $|R_t(a,0)|\le K_B(e_t+h_t)$ for a constant $K_B$
depending only on $D_{\mathcal A}$, the Lipschitz constants, $M_B$,
and $N$.
\end{lemma}

Summing the inequality of Lemma~\ref{lem:drift} along
$(a,\nu)=(a^\star,\mathbf0)$ telescopes the two potentials and
yields the static regret bound, with the step sizes balancing the
initial-distance term against the accumulated dual mismatch
$\alpha\beta T^2$. With the condition that prevents upper
clipping from discarding positive constraint increments, the same
one-step inequality further yields the cumulative constraint
violation bound.

\begin{theorem}[Static regret]\label{thm:regret}
Suppose Assumption~\ref{ass:regularity} holds,
$B=\Lambda_{\max}>0$ is fixed independently of $T$, and the step sizes follow the stated rules, then for every $T\ge1$,
\[
\E[\operatorname{Reg}(T)]\le C_{\rm reg}T^{2/3},
\]
where $C_{\rm reg}$ is independent of $T$ and given explicitly in
Appendix.
\end{theorem}

\begin{theorem}[Cumulative constraint violation]\label{thm:ccv}
Suppose the conditions of Theorem~\ref{thm:regret} hold and the
box satisfies $B>\sqrt{NH}$, where the constant $H$ is given
explicitly in the Appendix. Then for every $k=1,\ldots,m$ and
$T\ge1$,
\[
\E[\operatorname{CCV}_k(T)]\le C_{\rm ccv}T^{2/3},
\]
where $C_{\rm ccv}$ is independent of $T$.
\end{theorem}

\section{Numerical Simulation}\label{sec:simulation}
We evaluate the mechanism on an online distributed quadratic program
with dense coupling. A network of $N$ agents, each holding
$a_{i,t}\in[0,1]^{d_i}$, faces the stage cost
\begin{equation*}
    f_t(a)
    =
    \tfrac12\|Aa-b_t\|^{2}
    +
    \tfrac{\eta}{2}\|a-u_t\|^{2},
    \
    g_t(a)=\langle\mathbf 1,a\rangle-h\preceq\mathbf0,
    \label{eq:sim_problem}
\end{equation*}
where $A\in\mathbb R^{d\times d}$, $d=\sum_i d_i$, is a fixed dense
matrix with entries drawn from $\mathcal N(0,1/d)$, so that neither
the cost nor the constraint is separable across agents. 
We set $b_t=\bar b+0.05\sin(t/300)+0.02\,\zeta_t$ with $\zeta_t\sim\mathcal N(0,I)$, and let $u_t$ follow an AR(1) process with weight $\eta=0.01$.
The benchmark $a^\star$ is computed in closed form from the
realized sequence.

Agents run our proposed algorithm  over a ring
graph with Metropolis weights, with horizon-dependent constant
step sizes $\alpha_t=0.5T^{-1/2}$,
$\beta_t=T^{-1/2}$. We set
$T=3000$, link function $\varphi(u)=u/(1+u)$, and floor
$p_{\min}=0.15$. We compare three schemes under the
\emph{same communication expense} (the number of activated channels over the horizon): the proposed perturbation-aware algorithm; the prior work using full-communication belief consensus (DOPBC) in \cite{pan2026distributed} with
periodic activation, which transmits every $\lceil1/\rho\rceil$-th
round; and DOPBC with uniform random activation probability. Suppose the proposed algorithm activates channel $n_T$ times in total over the horizon $T$. We set $\rho = n_T/NT$, so that the periodic baseline transmits every $\lceil 1/\rho \rceil$-th round and the random baseline activates with probability $\rho$, ensuring all schemes share the same expected communication expense. Numerical results indicate that allocating communication via KKT perturbation, rather than via state variation, becomes increasingly effective in controlling regret as the belief dimension grows.
\begin{figure}[!t]
    \centering
    \includegraphics[width=\linewidth]{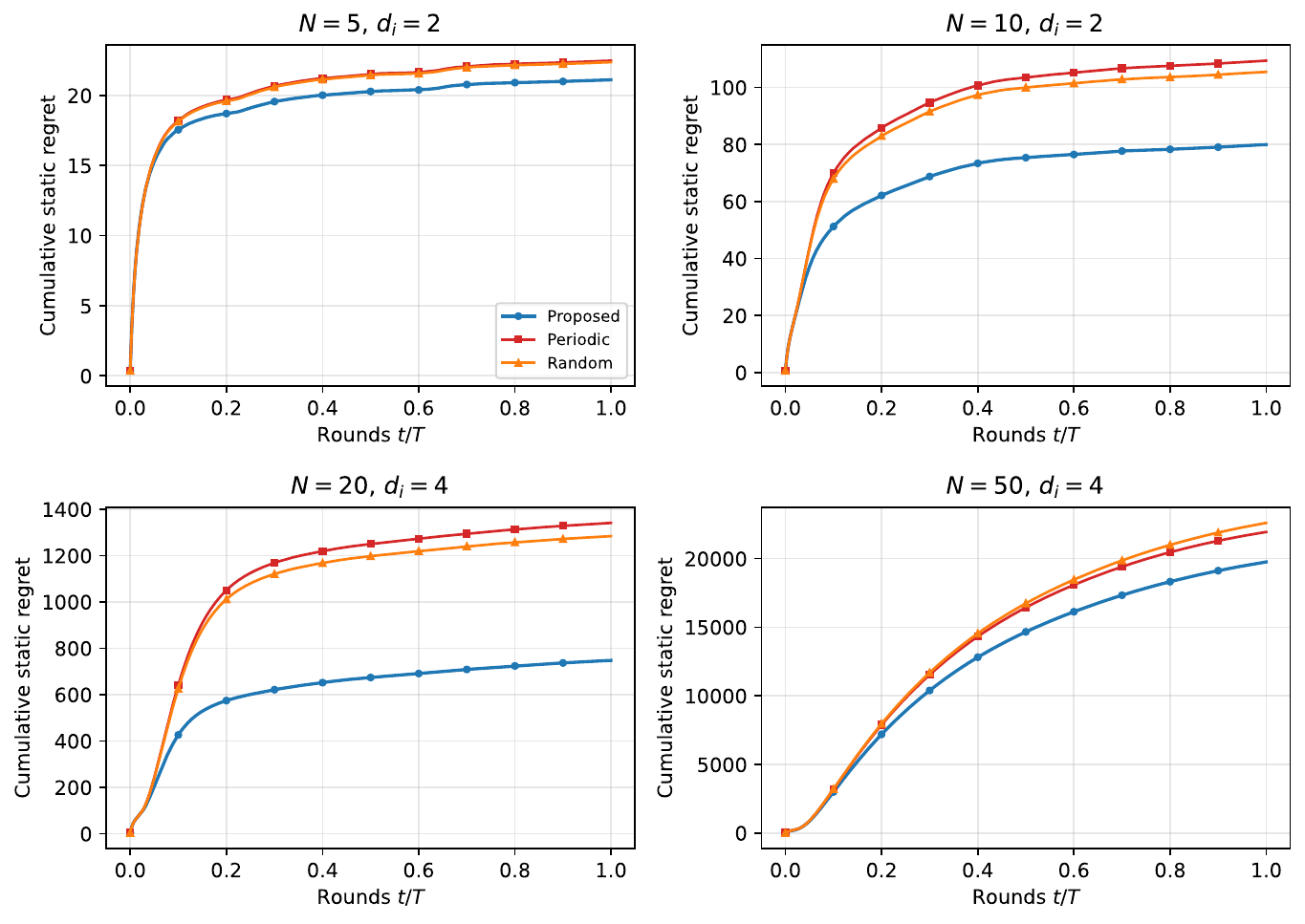}
    \caption{Cumulative static regret of the proposed KKT-perturbation
    triggering rule and  belief consensus in \cite{pan2026distributed} with periodic and random channel
    activation, under the same communication expense, i.e., the number of total activated channels over the horizon, for
    $(N,d_i)=(5,2),(10,2),(20,4),(50,4)$. }
    \label{fig:sim}
\end{figure}


\section{Conclusion}\label{sec:conclusion}
This paper addressed the communication overhead of belief-consensus
algorithms for distributed online convex optimization with nonseparable
costs and coupled constraints. We
proposed a randomized KKT-perturbation-aware communication mechanism
in which each primal-belief channel is activated with a probability
adapted to the channel impact on the local KKT operator, estimated causally from the most
recently revealed feedback, while the dual channel is augmented with
deterministic overrides that preserve the sign structure of its
staleness. Under standard regularity and a uniform Slater condition,
the algorithm achieves $O({T}^{2/3})$ expected static regret and
$O({T}^{2/3})$ expected cumulative constraint violation. We further showed that the triggering rule preserves the
channel-priority ordering induced by an idealized
communication-allocation problem, inspired by adaptive sampling in coordinate descent.

Several directions remain open for future extension. First, the idealized allocation
problem suggests enforcing the communication budget explicitly and
tracking the associated price in a decentralized manner, which would
upgrade the present priority principle into an online decentralized communication resource
allocation mechanism. Second, the triggering mechanism
developed in this work does not concern the payload, and hence combining it
with compressed message sharing would further reduce communication. Third, the bandit-feedback extension, directed
communication graphs, and dynamic-regret guarantees under
time-varying settings are worth investigating.

\bibliographystyle{IEEEtran}
\bibliography{ref}

@ARTICLE{pavel20distributed-gne,
  author={Pavel, Lacra},
  journal={IEEE Transactions on Automatic Control}, 
  title={Distributed GNE Seeking Under Partial-Decision Information Over Networks via a Doubly-Augmented Operator Splitting Approach}, 
  year={2020},
  volume={65},
  number={4},
  pages={1584-1597},
  doi={10.1109/TAC.2019.2922953}}

@article{liao2026fully, 
  year     = {2026}, 
  title    = {Fully Distributed {PrimalDual} Generalized Nash Equilibrium Seeking Algorithm Under Partial-Decision Information}, 
  author   = {Liao, Xinming and Meng, Wei and Li, Xiuxian}, 
  journal  = {{IEEE} Transactions on Control of Network Systems}, 
  issn     = {2325-5870}, 
  doi      = {10.1109/tcns.2025.3649096}, 
  pages    = {497--508}, 
  number   = {1}, 
  volume   = {13}
}

@ARTICLE{liu23game-erm,
  author={Liu, Shutian and Li, Tao and Zhu, Quanyan},
  journal={IEEE Transactions on Signal and Information Processing over Networks}, 
  title={Game-Theoretic Distributed Empirical Risk Minimization With Strategic Network Design}, 
  year={2023},
  volume={9},
  number={},
  pages={542-556},
  doi={10.1109/TSIPN.2023.3306106}}

@inproceedings{sinha24opt-coco, 
  year      = {2024}, 
  author    = {Sinha, Abhishek and Vaze, Rahul}, 
  title     = {Optimal Algorithms for Online Convex Optimization with Adversarial Constraints}, 
  booktitle = {Advances in Neural Information Processing Systems}, 
  doi       = {10.52202/079017-1306}, 
  pages     = {41274--41302}
}

@article{yuan20ma-coco, 
  year    = {2020}, 
  title   = {Distributed Online Optimization With Long-Term Constraints}, 
  author  = {Yuan, Deming and Proutiere, Alexandre and Shi, Guodong}, 
  journal = {{IEEE} Transactions on Automatic Control}, 
  issn    = {0018-9286}, 
  doi     = {10.1109/tac.2021.3057601}, 
  pages   = {1089--1104}, 
  number  = {3}, 
  volume  = {67}
}

@article{tao22confluence, 
  year    = {2022}, 
  title   = {The Confluence of Networks, Games, and Learning a Game-Theoretic Framework for Multiagent Decision Making Over Networks}, 
  author  = {Li, Tao and Peng, Guanze and Zhu, Quanyan and Ba\c{s}ar, Tamer}, 
  journal = {{IEEE} Control Systems}, 
  issn    = {1066-033X}, 
  doi     = {10.1109/mcs.2022.3171478}, 
  pages   = {35--67}, 
  number  = {4}, 
  volume  = {42}
}

@article{yi19doco, 
  year     = {2020}, 
  title    = {Distributed Online Convex Optimization With Time-Varying Coupled Inequality Constraints}, 
  author   = {Yi, Xinlei and Li, Xiuxian and Xie, Lihua and Johansson, Karl H.}, 
  journal  = {{IEEE} Transactions on Signal Processing}, 
  issn     = {1053-587X}, 
  doi      = {10.1109/tsp.2020.2964200}, 
  pages    = {731--746}, 
  volume   = {68}
}

@article{yangtao20doco-bandit, 
  year     = {2020}, 
  title    = {Distributed Bandit Online Convex Optimization With Time-Varying Coupled Inequality Constraints}, 
  author   = {Yi, Xinlei and Li, Xiuxian and Yang, Tao and Xie, Lihua and Chai, Tianyou and Johansson, Karl Henrik}, 
  journal  = {{IEEE} Transactions on Automatic Control}, 
  issn     = {0018-9286}, 
  doi      = {10.1109/tac.2020.3030883}, 
  pages    = {4620--4635}, 
  number   = {10}, 
  volume   = {66}
}

@article{tao25dima, 
  year    = {2025}, 
  title   = {Digital Twin-Based Driver Risk-Aware Predictive Mobility Analytics for Real-Time Situational Awareness Through Cooperative Sensing}, 
  author  = {Li, Tao and Bian, Zilin and Lei, Haozhe and Zuo, Fan and Yang, Ya-Ting and Zhu, Quanyan and Li, Zhenning and Chen, Zhibin and Ozbay, Kaan}, 
  journal = {{IEEE} Transactions on Intelligent Transportation Systems}, 
  issn    = {1524-9050}, 
  doi     = {10.1109/tits.2025.3604569}, 
  pages   = {20071--20090}, 
  number  = {11}, 
  volume  = {26}
}

@article{tao24picol, 
  year    = {2024}, 
  title   = {Multi-level traffic-responsive tilt camera surveillance through predictive correlated online learning}, 
  author  = {Li, Tao and Bian, Zilin and Lei, Haozhe and Zuo, Fan and Yang, Ya-Ting and Zhu, Quanyan and Li, Zhenning and Ozbay, Kaan}, 
  journal = {Transportation Research Part C: Emerging Technologies}, 
  issn    = {0968-090X}, 
  doi     = {10.1016/j.trc.2024.104804}, 
  pages   = {104804}, 
  volume  = {167}
}

@inproceedings{Tao_blackwell, 
  year     = {2021}, 
  author   = {Li, Tao and Peng, Guanze and Zhu, Quanyan}, 
  title    = {Blackwell Online Learning for Markov Decision Processes}, 
  doi      = {10.1109/ciss50987.2021.9400319}, 
  pages    = {1--6}, 
  volume   = {00}, 
  series   = {2021 55th Annual Conference on Information Sciences and Systems ({CISS})}
}

@article{guanze21local-gne, 
  year     = {2021}, 
  title    = {Locally-Aware Constrained Games on Networks}, 
  author   = {Peng, Guanze and Li, Tao and Liu, Shutian and Chen, Juntao and Zhu, Quanyan}, 
  journal  = {2021 American Control Conference ({ACC})}, 
  doi      = {10.23919/acc50511.2021.9482895}, 
  pages    = {4606--4611}, 
  volume   = {00}
}

@inproceedings{pan-tao23delay, 
  year      = {2023}, 
  author    = {Pan, Yunian and Li, Tao and Zhu, Quanyan}, 
  title     = {Is Stochastic Mirror Descent Vulnerable to Adversarial Delay Attacks? A Traffic Assignment Resilience Study}, 
  booktitle = {2023 62nd {IEEE} Conference on Decision and Control ({CDC})}, 
  doi       = {10.1109/cdc49753.2023.10384003}, 
  pages     = {8328--8333}
}

@inproceedings{pan-tao22noneq, 
  year      = {2023}, 
  author    = {Pan, Yunian and Li, Tao and Zhu, Quanyan}, 
  title     = {On the Resilience of Traffic Networks under Non-Equilibrium Learning}, 
  booktitle = {2023 American Control Conference}, 
  doi       = {10.23919/acc55779.2023.10156139}, 
  pages     = {3484--3489}
}

@article{lina13state_game, 
  year    = {2013}, 
  title   = {Designing Games for Distributed Optimization}, 
  author  = {Li, Na and Marden, Jason R.}, 
  journal = {{IEEE} Journal of Selected Topics in Signal Processing}, 
  issn    = {1932-4553}, 
  doi     = {10.1109/jstsp.2013.2246511}, 
  pages   = {230--242}, 
  number  = {2}, 
  volume  = {7}
}

@article{raginsky16nonseparable, 
  year    = {2016}, 
  title   = {Coordinate Dual Averaging for Decentralized Online Optimization With Nonseparable Global Objectives}, 
  author  = {Lee, Soomin and Nedic, Angelia and Raginsky, Maxim}, 
  journal = {{IEEE} Transactions on Control of Network Systems}, 
  doi     = {10.1109/tcns.2016.2573639}, 
  eprint  = {1508.07933}, 
  pages   = {34--44}, 
  number  = {1}, 
  volume  = {5}
}

@article{lin2025analysis,
  title={An analysis tool for push-sum based distributed optimization},
  author={Lin, Yixuan and Zhu, Zeru and Liu, Ji},
  journal={IEEE Transactions on Automatic Control},
  volume={70},
  number={12},
  pages={8298-8305},
  year={2025}
}

@article{yang2026dynamic,
  title={Dynamic Event-Triggered Nash Equilibrium Seeking for First-Order Multi-Agent Systems in Unreliable Networks},
  author={Yang, Yanping and Zheng, Yanan and He, Wangli and Han, Qing-Long},
  journal={IEEE Transactions on Automation Science and Engineering},
  year={2026},
  publisher={IEEE}
}

@article{chen2025event,
  title={Event-triggered adaptive Nash equilibrium seeking algorithm over unbalanced digraph},
  author={Chen, Jianing and Li, Yuxuan and Xue, Yuhan and Qin, Sitian},
  journal={IEEE Transactions on Automatic Control},
  volume={70},
  number={8},
  pages={5507--5514},
  year={2025},
  publisher={IEEE}
}

@article{ding2024distributed,
  title={Distributed Nash equilibrium seeking in merely monotone games using an event-and switching-activated communication scheme},
  author={Ding, Lei and Ye, Maojiao and Han, Qing-Long and Jia, Guobiao},
  journal={IEEE Transactions on Automatic Control},
  volume={70},
  number={1},
  pages={471--478},
  year={2024},
  publisher={IEEE}
}

@article{liu2023communication,
  title={A communication-efficient decentralized Newton's method with provably faster convergence},
  author={Liu, Huikang and Zhang, Jiaojiao and So, Anthony Man-Cho and Ling, Qing},
  journal={IEEE Transactions on Signal and Information Processing over Networks},
  volume={9},
  pages={427--441},
  year={2023},
  publisher={IEEE}
}

@article{zhu2026distributed,
  title={Distributed Online Randomized Gradient-Free Optimization With Compressed Communication},
  author={Zhu, Longkang and Shi, Xinli and Xu, Xiangping and Cao, Jinde and Chen, Xiangyong},
  journal={IEEE Transactions on Cybernetics},
  year={2026},
  publisher={IEEE}
}

@article{song2022compressed,
  title={Compressed gradient tracking for decentralized optimization over general directed networks},
  author={Song, Zhuoqing and Shi, Lei and Pu, Shi and Yan, Ming},
  journal={IEEE Transactions on Signal Processing},
  volume={70},
  pages={1775--1787},
  year={2022},
  publisher={IEEE}
}

@inproceedings{perekrestenko2017faster,
  title={Faster coordinate descent via adaptive importance sampling},
  author={Perekrestenko, Dmytro and Cevher, Volkan and Jaggi, Martin},
  booktitle={Artificial Intelligence and Statistics},
  pages={869--877},
  year={2017},
  organization={PMLR}
}

@article{leluc2022sgd,
  title={SGD with coordinate sampling: Theory and practice},
  author={Leluc, R{\'e}mi and Portier, Fran{\c{c}}ois},
  journal={Journal of Machine Learning Research},
  volume={23},
  number={342},
  pages={1--47},
  year={2022}
}

@article{pan2026distributed, 
  year     = {2026}, 
  title    = {Distributed Online Convex Optimization With Nonseparable Costs and Constraints}, 
  author   = {Pan, Zhaoye and Lei, Haozhe and Zuo, Fan and Bian, Zilin and Li, Tao}, 
  journal  = {{IEEE} Control Systems Letters}, 
  issn     = {2475-1456}, 
  doi      = {10.1109/lcsys.2026.3696347}, 
  pages    = {391--396}, 
  volume   = {10}
}

@article{tao22info,
  year = {2022},
  title = {The role of information structures in game-theoretic multi-agent learning},
  author = {Li, Tao and Zhao, Yuhan and Zhu, Quanyan},
  journal = {Annual Reviews in Control},
  issn = {1367-5788},
  doi = {10.1016/j.arcontrol.2022.03.003},
  pages = {296--314},
  volume = {53},
}

\appendix
\subsection{Proof of Lemma~\ref{lem:staleness}}\label{app:proof_staleness}
Fix an action block $k\in\mathcal V$ and measure errors against
the current action $a_{k,t}$:
\[
b^k_{i,t}=\mathsf R_kz_{i,t}-a_{k,t},\quad i\ne k,
\qquad
c^k_{j,t}=\mathsf R_ks^z_{j,t}-a_{k,t},\quad j\in\mathcal V,
\]
with $b^k_{k,t}=0$ since $z^k_{k,t}=a_{k,t}$. When agent $j$
broadcasts, its cache error is replaced by its belief error, so
with $c^{k,+}_{j,t}=\xi^z_{j,t}b^k_{j,t}+(1-\xi^z_{j,t})c^k_{j,t}$
the exact recursions are
\begin{align}
b^k_{i,t+1}
 &=w_{ii}b^k_{i,t}+\sum_{j\ne i}w_{ij}c^{k,+}_{j,t}
      -\Delta a_{k,t},\quad i\ne k,
      \label{eq:aug_b}\\
c^k_{j,t+1}&=c^{k,+}_{j,t}-\Delta a_{k,t},\quad j\in\mathcal V.
      \label{eq:aug_c}
\end{align}
Stack the $N-1$ non-ego belief errors and the $N$ cache errors
into $u^k_t$. Then \eqref{eq:aug_b}--\eqref{eq:aug_c} give
\[
u^k_{t+1}=(M_k(\xi_t^z)\otimes I_{d_k})u^k_t
      -\mathbf 1_{2N-1}\otimes\Delta a_{k,t},
\]
where $M_k$ is nonnegative and substochastic, and the missing
row mass corresponds to the exact ego error $b^k_{k,t}=0$.

We claim that the homogeneous products contract over $N$-round
windows. Fix a positive-weight spanning tree rooted at $k$,
ordered so that every parent precedes its children, and
prescribe one broadcast per vertex in this order over $N$
consecutive rounds, leaving other activations unrestricted.
Round-by-round conditioning gives this pattern conditional
probability at least $p^N$, where $p=p_{\min}^z$. Starting all
scalar entries at one, the root's cache error is zeroed at its
broadcast. A permanent deficit $b$ in a parent's cache error
creates deficits of at least $w_{ij}b$ in each child's belief
error at subsequent updates, and the child's own broadcast
transfers the deficit to its cache, where it is retained.
Induction down the tree shows that after $N$ rounds every row
sum of the window product is at most $1-\vartheta$, where
$\vartheta>0$ is the minimum root-to-vertex product of tree
weights, minimized also over roots. All other activation
patterns have product norm at most one.

Hence, with $q^2=1-p^N\vartheta(2-\vartheta)<1$ and
$\Phi_k(t,s)=M_k(\xi_{t-1}^z)\cdots M_k(\xi_s^z)$, the window
estimate yields
\begin{equation*}
\E[\|\Phi_k(t,s)\|_\infty^2\mid\mathcal F_s^-]
   \le q^{2\lfloor(t-s)/N\rfloor}.
\end{equation*}

Let $\|\cdot\|_{\max}$
denote the maximum norm. Since $u^k_1=0$ and
$\Delta a_{k,s}$ is $\mathcal F_{s+1}^-$-measurable, unrolling
the recursion gives
\[
(\E\|u^k_t\|_{\max}^2)^{1/2}
 \le\sum_{s<t}q^{\lfloor(t-s-1)/N\rfloor}
      (\E\|\Delta a_{k,s}\|^2)^{1/2}.
\]
The kernel $K_\ell=q^{\lfloor\ell/N\rfloor}$ is summable with
$\sum_{\ell\ge0}K_\ell\le N/(1-q)$, and the $\ell_2$
convolution inequality gives
\[
\sum_{t=1}^T\sum_k\E\|u^k_t\|_{\max}^2
 \le\frac{N^2}{(1-q)^2}\sum_{s=1}^{T-1}\E\|\Delta a_s\|^2
 \le CT\alpha^2,
\]
The
conditioning is essential because the increment depends on past
activations and cannot be taken out of the expectation of the
future product.

Every non-ego block of $\widetilde z_{i,t}-a_t$ is a convex
combination of cache errors $c^k_{j,t}$, so
$e_t^2\le(N-1)\sum_k\|u^k_t\|_{\max}^2$, which proves the first
two claims.

Nonexpansiveness of the box projection gives
\[
\|\lambda_{i,t+1}-\lambda_{j,t+1}\|
 \le\|\lambda_{i,t}-\lambda_{j,t}\|
      +\beta L_g\|\widetilde z_{i,t}-\widetilde z_{j,t}\|.
\]
Unrolling from $\lambda_{i,1}=0$ and writing
$D_t^2=\sum_i\|\lambda_{i,t}-\bar\lambda_t\|^2
=(2N)^{-1}\sum_{i,j}\|\lambda_{i,t}-\lambda_{j,t}\|^2$, we use
$\sum_{i,j}\|\widetilde z_i-\widetilde z_j\|^2
 \le2N\sum_i\|\widetilde z_i-a_t\|^2$, where the inequality
holds because the empirical mean minimizes the sum of squared
distances. Cauchy--Schwarz in time then gives
\[
\max_{t\le T}D_t^2\le\beta^2L_g^2T\sum_{t=1}^Te_t^2,
\qquad
\E\max_{t\le T}D_t^2\le C\alpha^2\beta^2T^2 .
\]

For $c^\lambda_{i,t}=s^\lambda_{i,t}-\lambda_{i,t}$, the cache
update gives
\[
c^\lambda_{i,t+1}
 =(1-\xi^\lambda_{i,t})c^\lambda_{i,t}
      -(\lambda_{i,t+1}-\lambda_{i,t}).
\]
Squaring, applying Young's inequality with parameter
$p_\lambda/2$ to the cross term, where
$p_\lambda=p_{\min}^\lambda$, and using
$\|\lambda_{i,t+1}-\lambda_{i,t}\|\le\beta G_H$ gives
\[
\E\|c^\lambda_{i,t+1}\|^2
 \le(1-p_\lambda/2)\E\|c^\lambda_{i,t}\|^2
      +3\beta^2G_H^2/p_\lambda.
\]
The increment may depend on the same activation, and no
independence is used. Iterating from $c^\lambda_{i,1}=0$ yields
$\E\|c^\lambda_{i,t}\|^2\le6\beta^2G_H^2/p_\lambda^2$, hence
$\sum_t\E\|c^\lambda_t\|^2\le CT\beta^2$ for the stacked cache
error $c^\lambda_t=\operatorname{col}(c^\lambda_{i,t})_i$.

Let $W_o=W-\operatorname{diag}(w_{ii})$ and
$J=\mathbf1\mathbf1^\top/N$. Stacking the fused dual states
and subtracting $\mathbf1\otimes\bar\lambda_t$ gives
$\widetilde\lambda_t-\mathbf1\otimes\bar\lambda_t$ as
\[
((W-J)\otimes I_m)
      (\lambda_t-\mathbf1\otimes\bar\lambda_t)
 +(W_o\otimes I_m)
      \operatorname{col}((1-\xi^\lambda_{i,t})c^\lambda_{i,t})_i,
\]
where both mixing matrices have Euclidean norm at most one.
Therefore $h_t\le D_t+\|c^\lambda_t\|$, and
\begin{align*}
\sum_{t=1}^T\E h_t
 &\le\sum_{t=1}^T(\E D_t^2)^{1/2}
      +\sqrt{T}\Big(\sum_{t=1}^T\E\|c^\lambda_t\|^2\Big)^{1/2} \\
 &\le C(\alpha\beta T^2+T\beta),
\end{align*}
while
\(
\E\max_{t\le T}h_t^2
 \le2\E\max_{t\le T}D_t^2+2\sum_{t=1}^T\E\|c^\lambda_t\|^2
 \le C(\alpha^2\beta^2T^2+T\beta^2).
\)
\hfill\QED

\subsection{Proof of Lemma~\ref{lem:drift}}\label{app:proof_one_step}

For the primal step, $a_i\in\mathcal A_i$ and nonexpansiveness
of the projection give
$\|a_{i,t+1}-a_i\|^2\le\|a_{i,t}-a_i-\alpha d_{i,t}\|^2$.
The primal blocks are distinct coordinates of one decision, so
summing over $i$ and using $\sum_i\|d_{i,t}\|^2\le G_B^2$
gives
\[
P_{t+1}(a)
 \le P_t(a)-2\alpha\langle d_t,a_t-a\rangle
      +\alpha^2G_B^2 .
\]
The dual copies concern the same multiplier and are averaged
rather than summed. Nonexpansiveness of the box projection
gives
\[
V_{t+1}(\nu)
 \le V_t(\nu)
      +\frac{2\beta}{N}\sum_i
       \langle g_{i,t},\lambda_{i,t}-\nu\rangle
      +\beta^2G_H^2 .
\]

It remains to convert the update terms into Lagrangian
differences. With the centralized direction
$d_t^{'}=\nabla f_t(a_t)+\nabla g_t(a_t)^\top\bar\lambda_t$
and the perturbation $\zeta_t=d_t-d_t^{'}$, convexity of
$f_t$ and of each component of $g_t$ gives
\[
\mathcal L_t(a_t,\bar\lambda_t)-\mathcal L_t(a,\bar\lambda_t)
 \le\langle d_t^{'},a_t-a\rangle
 =\langle d_t,a_t-a\rangle+R_t^a(a),
\]
where $R_t^a(a)=\langle\zeta_t,a-a_t\rangle$. For the
constraint term, splitting $g_t(a_t)$ into the average
$\bar g_t=N^{-1}\sum_i g_{i,t}$ and the residual gives
\[
\langle g_t(a_t),\nu-\bar\lambda_t\rangle
 =\frac1N\sum_i\langle g_{i,t},\nu-\lambda_{i,t}\rangle
 +R_t^g(\nu),
\]
where $R_t^g(\nu)=N^{-1}\sum_i
\langle g_{i,t}-g_t(a_t),\lambda_{i,t}-\nu\rangle$.

To bound the remainder at $\nu=0$, expand the $i$th primal
block of $\zeta_t$ as
\begin{align*}
\zeta_{i,t}
={}&\nabla_i f_t(\widetilde z_{i,t})-\nabla_i f_t(a_t)\\
 &+[\nabla_i g_t(\widetilde z_{i,t})
     -\nabla_i g_t(a_t)]^\top\widetilde\lambda_{i,t}\\
 &+\nabla_i g_t(a_t)^\top
      (\widetilde\lambda_{i,t}-\bar\lambda_t).
\end{align*}
By the Lipschitz and Jacobian bounds and
$\|\widetilde\lambda_{i,t}\|\le M_B$,
\[
\|\zeta_t\|
 \le(L_f+L_{\nabla g}M_B)e_t+G_gh_t .
\]
Hence $|R_t^a(a)|\le D_{\mathcal A}\|\zeta_t\|$, and
$\|g_{i,t}-g_t(a_t)\|\le L_g\|\widetilde z_{i,t}-a_t\|$ with
$\|\lambda_{i,t}\|\le M_B$ gives, by Cauchy--Schwarz over
agents,
\[
|R_t^g(0)|
 \le\frac{L_gM_B}{N}\sum_i\|\widetilde z_{i,t}-a_t\|
 \le\frac{L_gM_B}{\sqrt N}e_t .
\]
Adding the two bounds proves the remainder bound with
$K_B=D_{\mathcal A}(L_f+L_{\nabla g}M_B+G_g)
+L_gM_B/\sqrt N$.
\hfill\QED

\subsection{Proof of Theorem~\ref{thm:regret}}\label{app:proof_regret}
Set $(a,\nu)=(a^\star,0)$ in one-step inequality of Lemma \ref{lem:drift}, 
the full left side is
$f_t(a_t)-f_t(a^\star)-\langle\bar\lambda_t,g_t(a^\star)\rangle$,
which is at least the regret increment because
$g_t(a^\star)\preceq0$ and $\bar\lambda_t\succeq0$.
Summing in time before taking expectations yields
\begin{align*}
\operatorname{Reg}(T)
 \le{}&\frac{P_1(a^\star)-P_{T+1}(a^\star)}{2\alpha}
       +\frac{V_1(0)-V_{T+1}(0)}{2\beta}\\
 &+\frac{T}{2}(\alpha G_B^2+\beta G_H^2)
       +K_B\sum_{t=1}^T(e_t+h_t).
\end{align*}
Here the terminal potentials are nonnegative and $V_1(0)=0$,
so the dual potential contributes no positive initial term.
Using $P_1(a^\star)\le D_{\mathcal A}^2$ and
Lemma~\ref{lem:staleness} gives
\begin{equation}
\E\operatorname{Reg}(T)
 \le\frac{D_{\mathcal A}^2}{2\alpha}
          +C(T\alpha+T\beta+\alpha\beta T^2).
\label{eq:regret_bound}
\end{equation}
Following the stepsize rule, the initial-distance and accumulated
dual-mismatch terms are $O(T^{2/3})$, whereas $T\alpha$ and
$T\beta$ are $O(T^{1/3})$, which proves the claim.
\hfill\QED

\subsection{Proof of Theorem~\ref{thm:ccv}}\label{app:proof_ccv}
Let $\mathcal G_T=\{\max_{t\le T}(e_t+h_t)\le\kappa/(2K_B)\}$.
The deterministic bound
\[
\max_{t\le T}(e_t+h_t)^2
 \le2\sum_{t=1}^T e_t^2+2\max_{t\le T}h_t^2,
\]
followed by Markov's inequality
at threshold $\kappa^2/(4K_B^2)$, we have
$\mathbb P(\mathcal G_T^c)
 \le C(T\alpha^2+T\beta^2+\alpha^2\beta^2T^2).$
Consider $Q_t=V_t(0)+rP_t(\bar a)$, where $\bar a$ is the
Slater point. Set $(a,\nu)=(\bar a,0)$ in
one step inequality, using the Slater condition and
$f_t(\bar a)-f_t(a_t)\le F_0$ yield
\begin{align*}
Q_{t+1}-Q_t\le 2\beta\bigl[
 F_0-\varepsilon_{\rm S}\|\bar\lambda_t\|_1
 +(\alpha G_B^2+\beta G_H^2)/2\\
 {}+R_t^a(\bar a)+R_t^g(0)\bigr].
\end{align*}
For sufficiently large $T$, the stepsize terms are also at most
$\kappa/2$. Thus
\begin{equation}
Q_{t+1}-Q_t
 \le2\beta(F_0+\kappa-\varepsilon_{\rm S}\|\bar\lambda_t\|_1).
\label{eq:barrier_drift}
\end{equation}
Nonnegativity of every multiplier component gives
\[
N^2\|\bar\lambda_t\|_1^2
 =\Big(\sum_i\|\lambda_{i,t}\|_1\Big)^2
 \ge\sum_i\|\lambda_{i,t}\|^2=NV_t(0).
\]
Thus $\|\bar\lambda_t\|_1\ge\sqrt{V_t(0)/N}$.
If $Q_t\ge H$, then $V_t(0)\ge NR^2$, so the right side of
\eqref{eq:barrier_drift} is nonpositive. Below $H$, a one-step
increase is at most $2\beta(F_0+\kappa)$. Since
$Q_1\le rD_{\mathcal A}^2\le H$, induction gives
$Q_t\le H+2\beta(F_0+\kappa),\qquad t\le T+1.$
Consequently $\|\lambda_{i,t}\|\le\sqrt{NQ_t}$, for all sufficiently large $T$,
\[
\sqrt{NH+2N\beta(F_0+\kappa)}+\beta G_H<B.
\]
Every coordinate of the preprojection dual update is then below
$B$, which means no upper clipping occurs on $\mathcal G_T$.

Recall $\bar g_t=N^{-1}\sum_i g_{i,t}$.
Without upper clipping the dual projection only replaces negative
coordinates by zero. Therefore, 
$\lambda_{i,t+1}\succeq\lambda_{i,t}+\beta g_{i,t}$ on $\mathcal G_T$, we have
\[
\sum_{t=1}^T\bar g_t
 \preceq\frac{\bar\lambda_{T+1}-\bar\lambda_1}{\beta}
 =\frac{\bar\lambda_{T+1}}{\beta}.
\]
This is precisely the inequality that would fail if positive
increments were discarded at the upper boundary. Furthermore,
\[
\|g_t(a_t)-\bar g_t\|
 \le\frac{L_g}{N}\sum_i\|\widetilde z_{i,t}-a_t\|
 \le\frac{L_g}{\sqrt N}e_t.
\]
Combining these estimates componentwise gives
\[
\sum_{t=1}^T g_{t,k}(a_t)
 \le B/\beta+(L_g/\sqrt N)\sum_{t=1}^T e_t.
\]
The right side is nonnegative, so taking the positive part of
the cumulative sum proves the same bound for
$\operatorname{CCV}_k(T)$.
On $\mathcal G_T^c$ use the pathwise bound
$\operatorname{CCV}_k(T)\le G_HT$. Splitting the expectation,
\begin{align}
\E\operatorname{CCV}_k(T)
 &\le B/\beta+CT\alpha+G_HT\,\mathbb P(\mathcal G_T^c)\nonumber\\
 &\le B/\beta+CT\alpha\nonumber\\
 &\quad+C(T^2\alpha^2+T^2\beta^2+\alpha^2\beta^2T^3),
\label{eq:ccv_bound}
\end{align}
where $C$ is the constant depend on the fixed network, activation floors, regularity
constants, box, and stepsize coefficients.  Following the stepsize rules, $B/\beta=O(T^{2/3})$,
$T\alpha=O(T^{1/3})$, the first two exceptional-event terms
are $O(T^{2/3})$, and the last is $O(T^{1/3})$.
The thresholds used above are finite because $B-\sqrt{NH}>0$
and all constants are fixed independently of $T$.
\hfill\QED

\end{document}